%% file: Paper.tex
\documentclass{article}
\usepackage{ijcai22}

\usepackage{times}

\usepackage{soul}
\usepackage{url}
\usepackage[hidelinks]{hyperref}
\usepackage[utf8]{inputenc}
\usepackage[small, justification=centering, labelfont=bf]{caption}
\usepackage{graphicx}
\usepackage{wrapfig}
\usepackage{adjustbox}
\usepackage{float}
\usepackage{multirow}
\usepackage{amsmath}
\usepackage{booktabs}
\graphicspath{{Images/}}

\usepackage{tikz}
\usetikzlibrary{arrows,positioning}

\title{Boolean Decision Rules for Reinforcement Learning Policy Summarisation}
\author{
 James McCarthy$^{1,2}$\footnote{Contact Author}\and Rahul Nair$^1$\and Elizabeth Daly$^1$\and Radu Marinescu$^1$\and Ivana Dusparic$^2$\\
 \affiliations
 $^1$IBM Research Ireland\\
 $^2$Trinity College Dublin\\
 \emails
 james.mccarthy1@ibm.com, \{rahul.nair, radu.marinescu, elizabeth.daly\}@ie.ibm.com,\\ivana.dusparic@tcd.ie
}
\begin{document}

\maketitle

\begin{abstract}
Explainability of Reinforcement Learning (RL) policies remains a challenging research problem, particularly when considering RL in a safety context. Understanding the decisions and intentions of an RL policy offer avenues to incorporate safety into the policy by limiting undesirable actions. We propose the use of a Boolean Decision Rules model to create a post-hoc rule-based summary of an agent's policy. We evaluate our proposed approach using a DQN agent trained on an implementation of a lava gridworld and show that, given a hand-crafted feature representation of this gridworld, simple generalised rules can be created, giving a post-hoc explainable summary of the agent's policy.  We discuss possible avenues to introduce safety into a RL agent's policy by using rules generated by this rule-based model as constraints imposed on the agent's policy, as well as discuss how creating simple rule summaries of an agent's policy may help in the debugging process of RL agents.  
\end{abstract}

\section{Introduction}
\paragraph{}
While the introduction of deep learning techniques into Reinforcement learning (RL) research heralded in new possibilities and impressive progress, it also brought with it the difficulties inherent with deep learning \cite{glanois2021survey}. Challenges around the interpretability and explainability of these ever more complex techniques are just some of the issues facing RL and challenges the notion of Safety in RL. How confident can one be in a RL agent when they do not understand the reasoning behind its actions? If behaviour of RL agents can be captured in less complex, explainable rules humans can better identify and correct undesirable behaviour.

Generally in Machine Learning, and by extension RL, interpertability may, in some cases, result in a loss of predictive performance \cite{molnar2022}. In consideration of this potential loss of predictive performance, this work opts for a post-hoc explanation of the agent's learned policy rather than an interpretable policy. Similar work by \cite{AutonomousPolicyExplanation} looks at policy explanations by building rules consisting of predicates describing the state in a natural language form, interpretable to humans. A user can then query the agent regarding specific actions and receive an explanation of the state in which the action would occur. We believe that by creating a global rule-based policy summary, using features created directly from the environment, is a more robust policy summarisation method with regards to safety and potential use cases.
Other previous work such as \cite{Coppens2019DistillingDR}, focused on distilling the agent's policy into an interpretable decision tree and \cite{verma2018programmatically}, focused on programmatic interpretable policies, suffer performance drops due to adding interpretability. This trade-off between performance and interpretability is even more of a concern in a safety context. One might argue that if safety is a concern, performance should be a first priority. As such, when a hand-crafted feature set of the environment can be created, an assumption in our work, using this to create post-hoc explanations of an agent's policy may be more beneficial than creating an interpretable policy that relies on a hand-crafted feature set anyway. The Boolean Decision Rules (BDR) model presented by \cite{BooleanDecisionRules_dash_2018} generates rules that trade off between accurate classification and interpretability. It shows dominance in the accuracy-simplicity trade off and is competitive with other rule-learners (e.g. Decision Trees) whilst finding arguably simpler solutions \cite{BooleanDecisionRules_dash_2018}.

The main contribution of this paper is proposing the use of this BDR classification model to create rule-based summaries of an agents learned policy. We evaluate this model on a DQN policy trained in a safety-focused gridworld, creating a hand-crafted dataset by evaluating the agent in unseen environments. Using this dataset as input for the rules based model we show that simple explainable rules can be created summarising the agent's learned behaviour by predicting the agents next action, given the current state. We further discuss potential use cases of these rule-based policy summaries, such as feeding the rules back as constraints, and use cases around debugging the agent's policy.
\section{Methodology}\label{Methodology}
Given a trained agent, we gather sample trajectories of its interaction with an environment and record state and action pairs. In a supervised setting, we then seek to induce rules that describe specific actions based on state information.

\begin{figure}
\resizebox{0.95\columnwidth}{!}{
\input{figure}
}
\caption{Flow from training of agent to Decision Rule extraction. 1. Agent is trained in training environments. 2. The trained agent is evaluated in test environments to create a dataset of state-action pairs. 3. A BDR model is trained on this dataset to predict the agents action given the state. 4. Rules are then extracted from this model to summarise the agents policy.}
\end{figure}
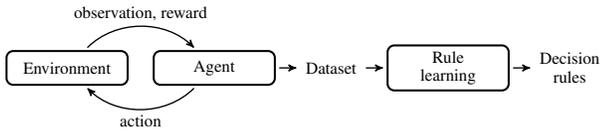
\begin{figure}
\centering
\includegraphics[width=0.35\textwidth]{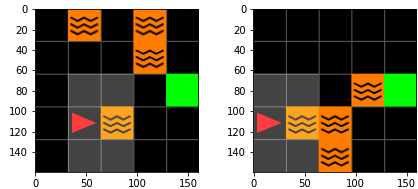}
\caption{Example of Lava Gridworld.\\ \textit{Black}: Empty squares, \textit{Orange}: Lava, \textit{Green}: Goal and \textit{Shaded} area of hand-crafted feature space}
\label{Gridworld}
\end{figure}

A Rainbow DQN \cite{RainbowCombiningImprovements_hessel_2018} agent is trained in a variation of the environment presented in \cite{AISafetyGridworlds_leike_2017}, \cite{ConciousnessBasedPlanning}. It is a fully observable 5x5 gridworld. The task is to reach the goal block whilst avoiding the lava blocks, which terminate the episode, see Figure \ref{Gridworld} for a visual example. The reward structure proposed in \cite{AISafetyGridworlds_leike_2017} is modified in two ways. The optimal reward for the environment is +1.9 instead of 1. +1 for reaching the reward and +0.9 which is scaled by the number of steps the agent takes with respect to the shortest path from its starting position to the goal. Each step remains a reward of 0, and stepping into lava is changed from a reward of 0 to -1, in order to punish stepping into lava. The agents policy was represented by a Convolutional Neural Network trained on the fully observable pixel representation of the gridworld. \footnote{Further details on experimental parameters for DQN agent and CNN are included in supplementary material}

A simple 3x2 grid in front of the agent is created by directly accessing the environment's class, shown as the shaded area in Figure \ref{Gridworld}. This grid details what is in the each block, i.e. Lava, Wall, Goal or Empty. The intention is to limit hand crafting of features other than simply recording what is in each block of the grid. A dataset of these features is created by evaluating the policy on a test set of 500 environments and capturing the state-action pairs at each timestep. 

A BDR model is created using the AIX360 toolkit\footnote{https://github.com/Trusted-AI/AIX360} and trained on the dataset of state-action pairs to classify the action the agent will take at each timestep, given the state. Because the model is a binary classification model, the classification is done in two stages. The first stage classifies the action as either Forward or Turn. Then the second stage classifies the turning actions as either Left or Right. To classify actions the BDR model creates rule sets in disjunctive normal form (DNF) i.e. rules sets consisting of OR-of-ANDs, whilst balancing an accuracy-simplicity trade off \cite{BooleanDecisionRules_dash_2018}. An example, in the context of this work would be ``IF (forward == lava) OR (forward == wall AND left == empty) THEN action==TURN". Rules are extracted from the trained models to create the summaries of the agent's learned policy.
 
\section{Results}
\paragraph{}
The DQN agent achieves close to the optimal result of 1.9. Results are averaged over 3 training runs set to three different seeds, shown in Table \ref{tab:AgentResults}. 
There are rare occasions when the agent fails to reach the goal, failing in two different modes. Either by stepping into the lava or by failing to reach the goal before the maximum number of steps is reached in the environment. In the latter case the agent gets stuck in some position in the gridworld and fails to get out of it. 
The BDR model shows good accuracy in predicting the agent's next action given its current state. The BDR model is trained on a dataset of state-action pairs created across 1500 evaluation episodes, 500 episodes from each of the 3 seeds. Table \ref{tab:AgentResults} shows the accuracy and F1-scores of the two binary classification stages, Forward/Turn and Left/Right. The high accuracy indicates that the agent is following consistent rules in its actions and that BDR model was able to extract them. The rules generated, describing the agent's behaviour, by this model back this up. Table \ref{tab:AgentResults} shows these rules, along with the observed instances of agreement and disagreement.
\begin{table*}[t]
    \centering
    \begin{adjustbox}{width=\textwidth}
    \begin{tabular}{c|c|c|c|c|c|c|c}
    \hline
         \textbf{Mean Reward} & \textbf{Success Rate} & \textbf{Action} &\textbf{Rule accuracy} & \textbf{F1-score} & \textbf{Rule} &\textbf{Agreement} & \textbf{Disagreement} \\
         \hline
        \multirow{6}{*}{$1.83 \pm 0.37$} &\multirow{6}{*}{97.0\%}&\multirow{2}{*}{Turn}&
        \multirow{2}{*}{80\%}&\multirow{2}{*}{62\%}&forward == lava&0.99&0.01\\
         &&&&&forward == wall&0.58&0.42\\
         \cline{3-8}
         &&\multirow{4}{*}{Right}& \multirow{4}{*}{84\%}&\multirow{4}{*}{83\%}&left == wall&0.88&0.12\\
         &&&&&      right == goal&1&0\\
         &&&&&      forward\_right == lava AND forward == lava&0.85&0.15\\
         &&&&&      forward\_right == empty AND forward == lava AND left == Empty&0.69&0.31\\
         
    \hline
    \end{tabular}
    \end{adjustbox}
    \caption{Agent's and BDR performance metrics and created rules\\
    \textit{Mean Reward}: cumulative reward of the agent, averaged over all evaluation episodes. \textit{Success Rate}: percentage of successfully completed episodes. \textit{Action}: action predicted by the BDR model. \textit{Rule Accuracy} and \textit{F1-score}: accuracy and F1-score of BDR model in predicting actions. \textit{Rule}: rule generated by the BDR model to predict the action. \textit{Agreement} and \textit{Disagreement}: proportion of agreement/disagreement between the rule and the agents action.}
    \label{tab:AgentResults}
\end{table*}

\section{Discussion}
\paragraph{}
The rules created by our approach give a global summary of the agent's learned policy. Rules indicating the agent will turn if ``forward == lava'' or ``forward == wall'' are examples of clear, intuitive rules that the agent should always follow to avoid states where catastrophic failure or wasteful actions occur. We outline several avenues where such rule-based summaries of policies are useful. 

\emph{Guardrails:} Rules like this can help give examples of state-action pairs that lead to states we wish to prevent the agent from ever reaching.  Despite the strong performance of the agent in terms of mean reward, there are still instances when it stepped into the lava or chose to move forward into the wall - the former being a catastrophic failure, the latter being a wasteful action. Unsafe behaviour like this could be avoided by feeding back these rules as constraints to the agent's policy. This could be done in two ways, either in a post-hoc manner after the agent has been trained or adding them in a more active learning and iterative process during training \cite{saunders2017trial}. While the former would require less human input and less altering of the agent's framework, it is arguably less safe and may only be useful in situations when new conditions of the environment have caused new unsafe states to arise. The latter, on the other hand, is a more involved process but may lead to more safety being incorporated to the agent's policy, if human feedback, by way of rule constraints, can push the policy towards more desirable actions during training. \cite{saunders2017trial} shows while in theory users-in-the-loop feedback can help avoid unsafe behaviour, the amount of feedback required to label instances does not scale. As a result, feedback based on rules can bring in user feedback while reducing the cognitive load on the user.

\emph{Debugging:} Rule-based summaries provide an interpretable view on the decision making logic of agents. This allows for identification of errant behaviours and determine how the agent is likely to generalise to novel environments. Previous work such as \cite{VisualizationDebuggingRL} suggests opting for visual debugging, but this can be problematic in giving a global summary of an agent's policy. Work creating interpretable policies, \cite{Coppens2019DistillingDR}, \cite{verma2018programmatically}, add similar debugging avenues to post-hoc rule-based summaries, but confine the policy to that underlying interpretable model. Whereas post-hoc summaries are agnostic to the underlying representation of the agent's policy, thereby maintaining the highest possible performance of an agent

\emph{Elicit human feedback:} Rule summaries can be used to gather human feedback on several aspects on policies. Rules can provide guidance on shaping reward functions, provide transparency on the underlying decision logic of the agent, and confirm that agents behave as intended. 

\section{Future Work}
\paragraph{}
This work has shown that given a hand-crafted feature set of a gridworld environment a set of explainable rules can be created, by way of a BDR model, describing an agents learned policy. While this is a relatively simple environment it presents the idea and potential of post-hoc rule-based policy summaries. However, global policy summaries like this can sometimes miss the nuance of particular decisions. Careful sampling of trajectories could address some challenges around this. Methods aimed at generating interpretable summaries for more complex tasks and environments would need additional considerations. 
A trade-off between the complexity of a policy summary and the amount of information a human user can process, will always remain. Scalable rule-based policy summaries that remain simple for human users to process may provide a middle solution.
\bibliography{ijcai22}
\bibliographystyle{named}
\pagebreak
\appendix
\onecolumn
\setcounter{table}{1}
\section{Supplementary Materials}
\subsection{Agent Parameters}
The Rainbow DQN agent was implemented using the Tianshou RL python package\footnote{https://github.com/thu-ml/tianshou}. Hyperparameters were tuned over 100 runs, for a maximum of 2,000,000 time steps. Early stopping was applied after 10 epochs if the evaluation reward did not increase by 0.01. These hyperparameter tuning runs were implemented using the MLFlow python package \footnote{https://github.com/mlflow/mlflow}. These hyperparameters are shown in Table \ref{tab:DQN Hyp}
\begin{table}[H]
    \renewcommand\thetable{A.1}
    \centering
    \begin{adjustbox}{}
    \begin{tabular}{|c|c|}
    \hline
         \textbf{Hyperparemeter} & \textbf{Value} \\
         \hline
         Atoms & 13\\
         Noisy Linear Layers& True\\
         Noise std & 0.1 \\
         Duelling & True \\
         Priority Replay Buffer & True \\
         Buffer Alpha & 0.4955\\
         Learning Rate & 0.00183 \\ 
         n-step &1\\
    \hline
    \end{tabular}
    \end{adjustbox}
    \caption{Hyperparameters for Rainbow DQN Agent}
    \label{tab:DQN Hyp}
\end{table}
\subsection{Convolutional Neural Network Parameters}
A 3-layer CNN was used in this work and was not part of any hyperparameter tuning and remained constant throughout. It was implemented using the PyTorch library\footnote{https://github.com/pytorch/pytorch}.  The pixel images of the environment were resized to a 40x40x3 pixel space. The hyperparemeters of the CNN are shown in table \ref{tab:CNNHyp}
\begin{table}[H]
    \renewcommand\thetable{A.2}
    \centering
    \begin{adjustbox}{}
    \begin{tabular}{|c|c|}
    \hline
         \textbf{Hyperparemeter} & \textbf{Value} \\
         \hline
         Convolution filters & [16, 32, 64]\\
         Kernel Size & [(4,4), (4,4), (3,3)]\\
         Stride & [2, 2, 1]\\
         Activation Function & ReLu\\
    \hline
    \end{tabular}
    \end{adjustbox}
    \caption{Hyperparameters for CNN}
    \label{tab:CNNHyp}
\end{table}
\subsubsection{Gridworld Implementation}
The gridworld implementation built off the MiniGrid python package \footnote{https://github.com/maximecb/gym-minigrid} and followed closely the work laid out by \cite{ConciousnessBasedPlanning} and their public github repo\footnote{https://github.com/mila-iqia/Conscious-Planning}. The randomisation's of each grid were ensured to by utilising the hash functionality of the MiniGrid package and building in a check to make sure each environment's hash was unique.
\end{document}

%% file: figure.tex
\begin{tikzpicture}[node distance=0.5cm, auto,
>=stealth',
    punkt/.style={
           rectangle,
           rounded corners,
           draw=black, very thick,
           text width=6.5em,
           minimum height=2em,
           text centered},
    pil/.style={
           ->,
           thick,
           shorten <=2pt,
           shorten >=2pt,}]
 \node[punkt] (env) {Environment};
 \node[punkt,right=of env] (agent) {Agent};
 
\path[pil, bend left=45] (env) edge node {observation, reward} (agent);
\path[pil, bend left=45] (agent) edge node {action} (env);

\node[right=of agent] (data) {Dataset}
    edge[pil, <-] (agent);
\node[punkt, right=of data, align=center] (rule) {Rule \\learning}
    edge[pil, <-] (data);
\node[right=of rule, align=center] (output) {Decision \\ rules}
    edge[pil, <-] (rule);
\end{tikzpicture}

%% file: Paper.bbl
\begin{thebibliography}{}

\bibitem[\protect\citeauthoryear{Coppens \bgroup \em et al.\egroup
  }{2019}]{Coppens2019DistillingDR}
Youri Coppens, Kyriakos Efthymiadis, Tom Lenaerts, and Ann Now{\'e}.
\newblock Distilling deep reinforcement learning policies in soft decision
  trees.
\newblock In {\em IJCAI 2019}, 2019.

\bibitem[\protect\citeauthoryear{Dash \bgroup \em et al.\egroup
  }{2018}]{BooleanDecisionRules_dash_2018}
Sanjeeb Dash, Oktay Gunluk, and Dennis Wei.
\newblock Boolean {{Decision Rules}} via {{Column Generation}}.
\newblock In {\em Advances in {{Neural Information Processing Systems}}},
  volume~31. {Curran Associates, Inc.}, 2018.

\bibitem[\protect\citeauthoryear{Deshpande \bgroup \em et al.\egroup
  }{2020}]{VisualizationDebuggingRL}
Shuby Deshpande, Benjamin Eysenbach, and Jeff Schneider.
\newblock Interactive visualization for debugging rl, 2020.

\bibitem[\protect\citeauthoryear{Glanois \bgroup \em et al.\egroup
  }{2021}]{glanois2021survey}
Claire Glanois, Paul Weng, Matthieu Zimmer, Dong Li, Tianpei Yang, Jianye Hao,
  and Wulong Liu.
\newblock A survey on interpretable reinforcement learning.
\newblock {\em arXiv preprint arXiv:2112.13112}, 2021.

\bibitem[\protect\citeauthoryear{Hayes and
  Shah}{2017}]{AutonomousPolicyExplanation}
Bradley Hayes and Julie~A. Shah.
\newblock Improving robot controller transparency through autonomous policy
  explanation.
\newblock In {\em Proceedings of the 2017 ACM/IEEE International Conference on
  Human-Robot Interaction}, HRI '17, page 303–312, New York, NY, USA, 2017.
  Association for Computing Machinery.

\bibitem[\protect\citeauthoryear{Hessel \bgroup \em et al.\egroup
  }{2018}]{RainbowCombiningImprovements_hessel_2018}
Matteo Hessel, Joseph Modayil, Hado {van Hasselt}, Tom Schaul, Georg Ostrovski,
  Will Dabney, Dan Horgan, Bilal Piot, Mohammad Azar, and David Silver.
\newblock Rainbow: {{Combining Improvements}} in {{Deep Reinforcement
  Learning}}.
\newblock {\em Proceedings of the AAAI Conference on Artificial Intelligence},
  32(1), April 2018.

\bibitem[\protect\citeauthoryear{Leike \bgroup \em et al.\egroup
  }{2017}]{AISafetyGridworlds_leike_2017}
Jan Leike, Miljan Martic, Victoria Krakovna, Pedro~A. Ortega, Tom Everitt,
  Andrew Lefrancq, Laurent Orseau, and Shane Legg.
\newblock {{AI Safety Gridworlds}}.
\newblock {\em arXiv:1711.09883 [cs]}, November 2017.

\bibitem[\protect\citeauthoryear{Molnar}{2022}]{molnar2022}
Christoph Molnar.
\newblock {\em Interpretable Machine Learning}.
\newblock 2 edition, 2022.

\bibitem[\protect\citeauthoryear{Saunders \bgroup \em et al.\egroup
  }{2017}]{saunders2017trial}
William Saunders, Girish Sastry, Andreas Stuhlmueller, and Owain Evans.
\newblock Trial without error: Towards safe reinforcement learning via human
  intervention.
\newblock {\em arXiv preprint arXiv:1707.05173}, 2017.

\bibitem[\protect\citeauthoryear{Verma \bgroup \em et al.\egroup
  }{2018}]{verma2018programmatically}
Abhinav Verma, Vijayaraghavan Murali, Rishabh Singh, Pushmeet Kohli, and Swarat
  Chaudhuri.
\newblock Programmatically interpretable reinforcement learning.
\newblock In {\em International Conference on Machine Learning}, pages
  5045--5054. PMLR, 2018.

\bibitem[\protect\citeauthoryear{Zhao \bgroup \em et al.\egroup
  }{2021}]{ConciousnessBasedPlanning}
Mingde Zhao, Zhen Liu, Sitao Luan, Shuyuan Zhang, Doina Precup, and Yoshua
  Bengio.
\newblock A consciousness-inspired planning agent for model-based reinforcement
  learning.
\newblock In M.~Ranzato, A.~Beygelzimer, Y.~Dauphin, P.S. Liang, and J.~Wortman
  Vaughan, editors, {\em Advances in Neural Information Processing Systems},
  volume~34, pages 1569--1581. Curran Associates, Inc., 2021.

\end{thebibliography}
